\documentclass{article}
\usepackage[utf8]{inputenc}
\usepackage[T1]{fontenc}

\usepackage[margin=1in, columnsep=15pt, includehead, headheight=1.5cm, includefoot]{geometry}
\usepackage[numbers, square, compress]{natbib}
\usepackage{microtype} 
\usepackage{graphicx}
\usepackage{subcaption}
\usepackage{booktabs} 
\usepackage{multicol} 
\usepackage{fancyhdr} 
\usepackage{marvosym} 
\usepackage{amsmath,amssymb,mathtools,amsthm}

\usepackage[table]{xcolor} 
\usepackage{float} 
\usepackage{multirow}      
\usepackage{xspace}
\usepackage{colortbl}
\usepackage{multirow}
\usepackage{ulem}
\usepackage{tcolorbox} 
\tcbuselibrary{skins} 
\usepackage{hyperref}
\hypersetup{
    colorlinks=true,
    linkcolor=blue,
    citecolor=blue,
    urlcolor=blue,
    breaklinks=true
}
\theoremstyle{plain}

\theoremstyle{definition}

\theoremstyle{remark}

\usepackage[disable,textsize=tiny]{todonotes}
\newcommand{\runningtitle}{OpenEAI-Platform}
\newcommand{\splatform}{\textit{OpenEAI-Platform}\xspace}
\newcommand{\svla}{\textit{OpenEAI-VLA}\xspace}
\newcommand{\sarm}{\textit{OpenEAI-Arm}\xspace}

\renewcommand{\headrulewidth}{0.4pt}
\fancypagestyle{titlepage}{
  \fancyhf{}
  \lhead{\includegraphics[height=2cm]{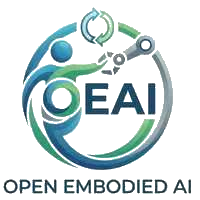}}
  \chead{\textbf{\runningtitle}}
  \rhead{2026.02.25}
  \renewcommand{\headrulewidth}{0.4pt}
  \fancyfoot[L]{\small \textsuperscript{*} Equal contribution. \quad   \textsuperscript{\Letter}  Corresponding author.} 
  \renewcommand{\footrulewidth}{0.4pt}
}
\fancypagestyle{main}{
  \fancyhf{}
  \lhead{\includegraphics[height=2cm]{imgs/OpenEAI.png}}
  \chead{\textbf{\runningtitle}}
  \rhead{2026.02.25}
  \renewcommand{\headrulewidth}{0.4pt}
  \fancyfoot[C]{\thepage}
  \renewcommand{\footrulewidth}{0.4pt}
}

\usepackage{etoolbox}
\patchcmd{\maketitle}{\thispagestyle{plain}}{\thispagestyle{titlepage}}{}{}

\makeatletter
\renewcommand{\@maketitle}{%
  \newpage
  \null
  \vskip 0em
  \begin{center}%
  \let \footnote \thanks
    {\LARGE\bfseries\boldmath \@title \par}
    \vskip 1.5em%
    {\large \@author \par}
    \vskip 1em%
  \end{center}%
  \par
  \vskip 1.5em}
\makeatother

\usepackage{epigraph}
\usepackage{enumitem}
\usepackage{cleveref}

\begin{document}


\pagenumbering{arabic}
\setcounter{page}{1}

\title{{OpenEAI-Platform: An Open-source Embodied Artificial Intelligence Hardware-Software Unified Platform}}


\author{
    {\bfseries Jinyuan Zhang}\textsuperscript{*,1,2}, 
    {\bfseries Luoyi Fan}\textsuperscript{*,1,3}, 
    {\bfseries Leiyu Wang}\textsuperscript{1,3}, 
    {\bfseries Yeqiang Wang}\textsuperscript{1,3}, \\
    {\bfseries Yicheng Zhu}\textsuperscript{4}, 
    {\bfseries Cewu Lu}\textsuperscript{1,3}, 
    {\bfseries Nanyang Ye}\textsuperscript{1,3,\href{mailto:ynylincoln@sjtu.edu.cn}\Letter} \\[0.3em]
    \textsuperscript{1}Shanghai Innovation Institute
    \textsuperscript{2}Huazhong University of Science and Technology; \\
    \textsuperscript{3}Shanghai Jiao Tong University; 
    \textsuperscript{4}Media
    \\[0.1em]
    Fully Open-Sourced at: \url{https://github.com/sii-research/ORoboSoul/blob/openeai-platform/}
}

\maketitle
\thispagestyle{titlepage}

\pagestyle{main}

\begin{figure*}[ht]
  \begin{center}
    \centerline{\includegraphics[width=\linewidth]{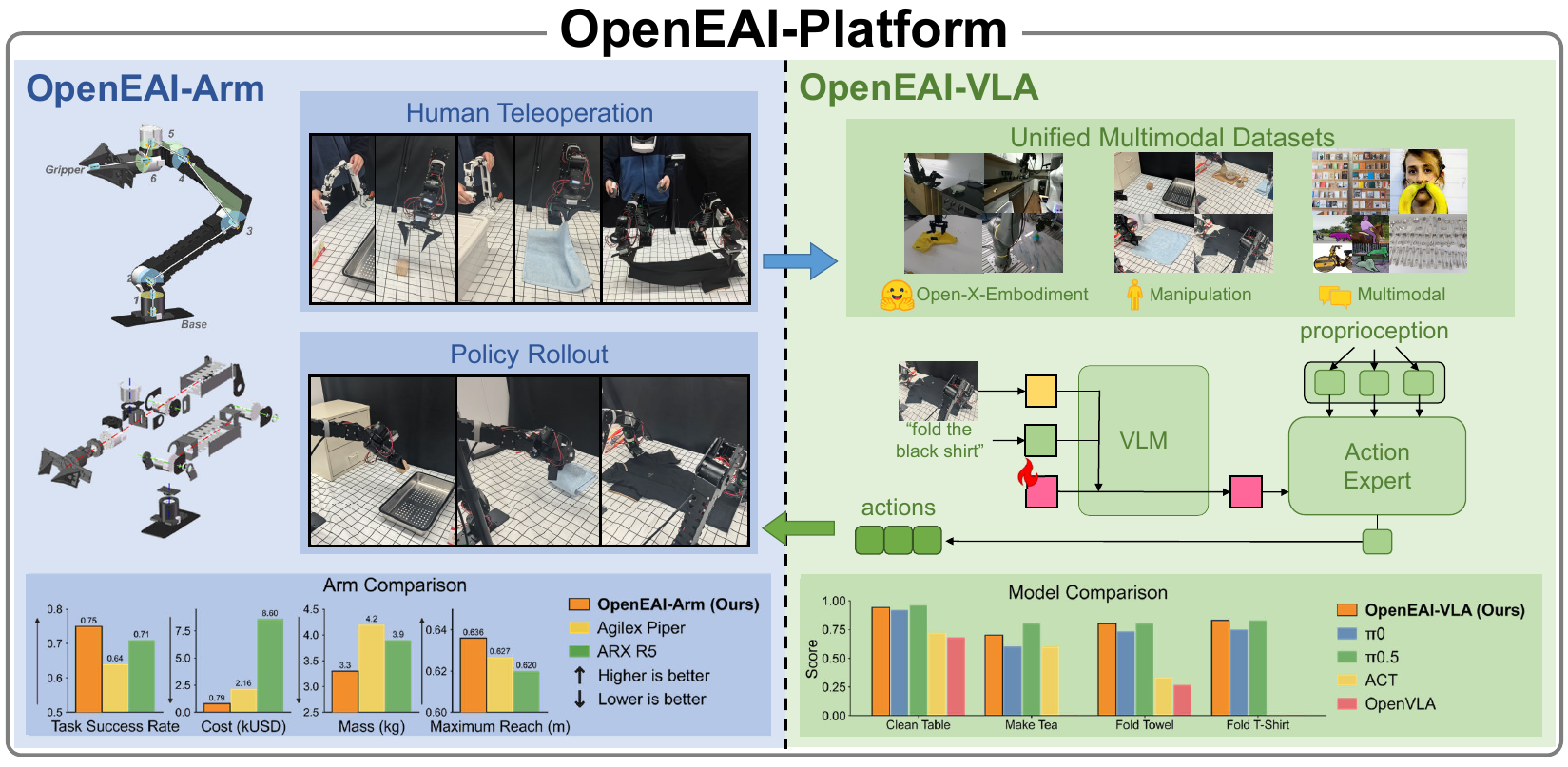}}
    \caption{
      Overall structure of \splatform. Left: \sarm enables low-cost data collection via human teleoperation and supports closed-loop policy rollouts on real manipulation tasks. Right: \svla is trained on unified multimodal datasets and maps visual observations, language instructions, and robot states to actions through a VLM backbone and an action expert. Bottom: quantitative comparisons for both the robotic arm and the learned policies across multiple tasks.
    }
    \label{fig:overall_architecture}
  \end{center}
  \vskip -0.2in
\end{figure*}

\begin{tcolorbox}[
    colback=blue!5, 
    colframe=white, 
    boxrule=0pt, 
    sharp corners, 
    parbox=false,
    boxsep=0.5em, 
    width=\linewidth 
]
\begin{center}
    \Large \textbf{Abstract}
\end{center}
\vspace{0in} 

Embodied AI in the real world requires both accurate hardware and robust vision-language-action (VLA) policies. We present \splatform, a fully open-source platform that integrates a low-cost 6+1 degree-of-freedom (dof) robotic arm (\sarm) and a reproducible VLA model (\svla). \sarm provides open-source mechanical designs for low manufacturing cost and compliant control methods for higher accuracy. \svla builds on Qwen3-VL-4B and uses a Diffusion Transformer action head, and is trained in two stages with only open-source robot and multimodal datasets. Across four real-world manipulation tasks, \sarm outperforms two commercial 6+1-dof arms under the same policy, and \svla achieves success rates comparable to the large-scale pretrained $\pi_0$ baseline with only limited pretraining data. We will release the full hardware designs, drivers, models, and training/data pipelines to support reproducible research and scalable data collection. Our codes, layouts, and models will be released after the paper is accepted.
\end{tcolorbox}

\vspace{0.2in} 


\begin{multicols}{2}[\setlength{\parskip}{-0.1pt}]

\section{Introduction}
\setlength{\epigraphwidth}{0.45\textwidth} 
\epigraph{Alone we can do so little; together we can do so much.}{--- \textit{Helen Keller}}

Embodied intelligence studies how intelligent behavior emerges from the tight coupling of an agent’s body, environment, and control algorithms, where hardware properties such as sensors, actuators, and mechanical structure are as critical as learned policies. In recent years, mainstream progress has been driven by policy learning, with Vision-Language-Action (VLA) models becoming the dominant approach; notably, the Physical Intelligence series ($\pi_0$ \cite{pi0}, $\pi_{0.5}$ \cite{pi0_5}) represents state-of-the-art performance but relies on large-scale proprietary datasets, creating a major reproducibility and data-scaling barrier. On the hardware side, many real-world VLA systems build on Stanford’s Mobile ALOHA teleoperation platform \cite{fu2024mobile} or commercial arms with similar 6 joints \(+\) 1 gripper degree-of-freedom (dof) manipulators, yet servos or platform-specific structural choices can limit workspace and payload, and differences across manipulators introduce cross-embodiment gaps that hinder transfer and fair comparison. Moreover, the high cost of commercial robotic arms restrict large-scale data collection and make it difficult to study the interaction between VLA algorithms and low-level control. Besides, the black-box nature of these robotic arms makes it challenging to investigate the complex coupling between VLA algorithms and low-level control methods. These pose significant bottlenecks for the advancement of the field. Together, these limitations form practical bottlenecks for scalable, reproducible progress in embodied manipulation.

To address reproducibility and scaling bottlenecks in real-world embodied manipulation, we introduce \splatform, a fully open-source platform that includes the low-cost 6 \(+\)1-DoF robotic arm \sarm with an end-to-end VLA policy \svla as illustrated in Figure \ref{fig:overall_architecture}. Our key insight is that progress in VLA systems depends on end-to-end reproducibility rather than model design alone, which requires public and standardized hardware specifications, manufacturing procedures, low-level drivers and controllers, data processing interfaces, and training recipes. In practice, many VLA systems remain difficult to reproduce because training infrastructure and private large-scale training datasets are not released and open-source robot datasets form ``data islands'' with inconsistent action/state definitions and formats. \splatform mitigates these issues by releasing the full pipeline and providing a unified interface for dataset integration and using only public datasets for pretraining and fine-tuning.

Concretely, we obtain \sarm by solving a two-objective MDH optimization problem that targets manipulation operability and energy efficiency under anthropometric constraints using NSGA-III. Reliable high-frequency execution of discrete VLA chunks is supported by a dynamics-compensated PID controller together with rolling action-chunking interpolation, which enforces smooth velocity and acceleration transitions under actuation limits. On the algorithmic side, \svla follows a two-stage training recipe using only open-source data, with broad pretraining on a curated mixture of public robot datasets and post-training on a small amount of in-domain demonstrations, optionally mixed with multimodal data. Cross-dataset training is enabled by a standardized dataset conversion and cleaning pipeline together with dataset-aware alignment modules that map heterogeneous state/action conventions into a shared interface. For perception-to-action learning, \svla couples a pretrained VLM backbone with a compact readout mechanism and a generative action head for continuous action-chunk prediction. We validate the complete stack on multiple real-world tasks, showing that \sarm achieves robust execution compared to commercial arms under the same policy, and that \svla trained purely on open-source datasets achieves competitive success rates relative to large-scale VLA baselines. To summarize, we make four contributions:
\begin{enumerate}
    \item A reproducible low-cost 6-DoF robotic arm with effective control algorithms, and its full manufacturing materials to support both replication and scalable production.
    \item A two-stage, fully reproducible data and training recipe, including a dataset conversion and cleaning pipeline and a curated open-source pretraining mixture together with a fine-tuning protocol that combines small in-domain demonstrations with multimodal data.
    \item An open-source VLA policy that couples a pretrained Qwen3-VL as backbone with a generative action head and can be trained and deployed through our released codebase from scratch.
    \item Comprehensive experiment validation for the complete system on multiple real-world manipulation tasks, showing strong hardware-level reliability and competitive success rates compared to recent large-scale SOTA VLA baselines and commercial robot platforms, despite relying only on open-source data and code.
\end{enumerate}

\section{Related Work}

\subsection{Robotic Arms}

There are many open-source and commercial robotic arms at present. The most popular open-source robotic arm platform is HuggingFace's LeRobot \cite{cadene2024lerobot}, which provides various kinds of robotics arms made with 3D-printed components and servo motors. However, most of them, including the latest SO-101 arm, have only 5 DoF with relatively short reach and low payload (estimated maximum reach 413.3 mm and payload 0.5 kg), and their repeatability is constrained by predominantly 3D-printed structures, making deployment in real scenarios challenging. Commercial robotic arms, like ARX R5\cite{arx_r5}, Realman Robotics RM65/RM75 series\cite{realman-rm}, Franka Robotics FR3\cite{franka-research3}, and Universal Robots UR5e\cite{ur5e}, typically provide 6-dof and 7-dof arms, whereas their high purchase cost, ranging from 7,000 USD to over 40,000 USD, limits large-scale deployment and reduces their accessibility in academic research and experimentation. Besides, their unique configurations also make it harder to realize cross-embodiment generalization for current models. More affordable 6-DoF arms, such as AgileX Piper \cite{agilex_piper} and I2RT YAM \cite{i2rt-yam} (approximately 1,500–3,000 USD), typically follow the ALOHA-style dual-arm teleoperation configuration with an ARX R5-like kinematic layout, with Piper differing mainly in its last three joints. This configuration has become a common baseline for desktop manipulation in Mobile-ALOHA and VLA-driven studies \cite{zhai2025igniting,fu2024mobile,pi0,pi0_5}. Despite their affordability, these arms are typically delivered as black-box platforms that expose only high-level interfaces, which limits embodiment-level customization of dynamic parameters and the actuation stack such as low-level torque control and torque feedback, increasing the difficulty of task-specific adaptation. In our design of \sarm, we adopt an ARX R5-like architecture while releasing all materials and low-level source code to enable customization, and we reduce the total manufacturing cost to approximately 790 USD to lower the barrier to large-scale deployment.


\subsection{Vision-Language-Action Models}

Recent robot manipulation policies focus on the design and improvement of VLA models, which mainly consist of a backbone VLM with an action head. The choices of backbone VLM include Qwen2-VL \cite{zhou2025chatvla2visionlanguageactionmodelopenworld, wen2025dexvla}, Qwen2.5-VL \cite{zhai2025igniting}, PaliGemma \cite{pi0, pi0_5}, Prismatic-VLM \cite{kim24openvla}, SmolVLM \cite{shukor2025smolvlavisionlanguageactionmodelaffordable}, and others. These VLMs varies from the smallest 500M SmolVLM \cite{marafioti2025smolvlm} to largest 7B Prismatic-VLM \cite{karamcheti2024prismaticvlmsinvestigatingdesign}, but most of them are within the size from 2B to 4B. Action heads are often contiguous policies like diffusion and flow matching \cite{chi2024diffusionpolicyvisuomotorpolicy, pi0, pi0_5} or discrete tokenization \cite{zhao2023learningfinegrainedbimanualmanipulation,kim24openvla}, and use conditioning embeddings from backbone VLM. Our \svla uses the latest Qwen3-VL-4B model \cite{Qwen3-VL} as our backbone VLM to utilize its strong vision-language alignment. For the action head, we use a multi-layer diffusion transformer as Diffusion Transformer Policy and some modern VLAs \cite{hou2025diffusiontransformerpolicy, wen2025dexvla} to decode the conditioning embeddings into velocity vectors for action generation via flow-matching like $\pi_0$\cite{pi0}.


Besides, many of the SOTA VLAs \cite{pi0,pi0_5,zhai2025igniting} only publish their fine-tuning and inference process, but keep pretraining codes and datasets private. Our \svla only uses the open-source Open-X-Embodiment dataset \cite{open_x_embodiment_rt_x_2023} as the only data source at pretraining stage, and uses COCO\cite{coco2017}, VQA-v2\cite{balancedvqav2} and pixmo-points\cite{pixmo} as the multi-modal datasets at fine-tuning stage. These datasets are all published and easy to access, and we provide full codes for the two training stages so that everyone can reproduce and improve the model from scratch. 

\begin{figure*}[t]
  \begin{center}
    \centerline{\includegraphics[width=\linewidth]{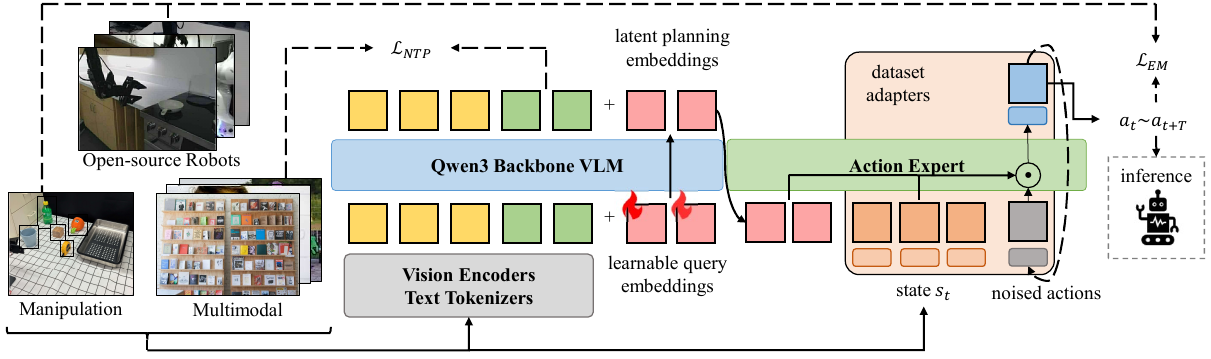}}
    \caption{
      Overview of our proposed framework. We leverage Qwen3 as the vision-language backbone, appending learnable query embeddings to extract latent planning representations from multimodal inputs. These representations, combined with proprioceptive state $s_t$, condition the diffusion-based Action Expert to generate action chunks via dataset-specific adapters. The model is jointly optimized using next-token prediction loss $\mathcal{L}_{NTP}$ for semantic understanding and action matching loss $\mathcal{L}_{EM}$ for policy learning.
    }
    \label{fig:vla_architecture}
  \end{center}
  \vskip -0.2in
\end{figure*}

\section{Policy Design}

To enable real-world task manipulation on our \sarm within \splatform, we develop \svla, an open-source VLA system and release the complete pipeline, including dataset conversion tools, pretraining/fine-tuning scripts, inference server interfaces, and model checkpoints. \svla targets two major barriers to reproducible and scalable VLA research: heterogeneous open-source datasets that form ``data islands'' with inconsistent state/action and sensor conventions, and strong VLA models that are difficult to reproduce due to non-public training details and infrastructure. Following the two-stage foundation-model recipe of recent VLAs \cite{pi0_5,octomodelteam2024octoopensourcegeneralistrobot,kim24openvla}, we pretrain on diverse public robot datasets and post-train on task demonstrations mixed with multimodal data. At the model level, we use a learnable query embedding to compress vision-and-instruction features into fixed-length conditioning and dataset-specific projectors to align cross-dataset state/action conventions into a shared space. Together with our unified preprocessing toolchain for converting heterogeneous datasets into a uniform format, \svla supports both large-scale pretraining for data-rich users and data-efficient fine-tuning for small-task users.

\paragraph{Model architecture.}
\svla uses Qwen3-VL-Instruct 4B as the backbone VLM, but instead of feeding all VLM hidden states to the action head, we introduce a fixed-length learnable feature query as a trainable readout interface between multimodal perception and action generation (illustrated in our architecture diagram). After tokenizing images and text into VLM input embeddings, we append a learnable token sequence $\mathbf{q}\in\mathbb{R}^{M\times d}$ of fixed length $M$ and the dimensionality of the VLM’s token embedding space $d$, and run the VLM on the concatenated sequence; we then extract only the final-layer hidden states corresponding to this query, $\mathbf{h}_q\in\mathbb{R}^{M\times d}$, as compact conditioning embeddings for action generation. This design provides a fixed-bandwidth bottleneck that stabilizes the VLM-to-action interface and keeps the action head’s compute largely independent of the number of image patches or text tokens, while allowing the query tokens to learn what task-relevant information to compress from vision and instruction inputs. Conditioned on $\mathbf{h}_q$, \svla generates continuous action chunks with a Diffusion-Transformer-style action head, following recent DiT-based robot policies \cite{hou2025diffusiontransformerpolicy}. our action expert is implemented as an 18-layer DiT-style decoder with 32 attention heads per layer. At each denoising step, the action head takes as input (i) a noisy action chunk, (ii) the current robot state, and (iii) the VLM-derived conditioning embeddings $\mathbf{h}_q$ from the learnable feature query, and outputs the velocity field $\mathbf{v}_\theta$ used by conditional flow matching for action denoising.

\paragraph{Two-stage training recipe.}
We train \svla with a two-stage recipe. In the pretraining stage, we use a weighted mixture of large open-source Open X-Embodiment subsets \cite{open_x_embodiment_rt_x_2023}, converted into our unified format with consistent preprocessing and filtering of incomplete or failed trajectories. The action expert is optimized with a $\pi_0$-style conditional flow matching objective \cite{pi0,lipman2023flowmatchinggenerativemodeling,liu2022rectifiedflowmarginalpreserving}: we construct noisy action chunks by interpolating between a clean target chunk and Gaussian noise, sample $\tau$ from a Beta distribution $p(\tau)$ emphasizing noisier timesteps \cite{pi0,shukor2025smolvlavisionlanguageactionmodelaffordable}, and train a vector-field network to match the denoising direction. The resulting embodied control loss for embodied data (EM) is
\begin{align}
\mathcal{L}_{\text{EM}}(\theta)
=
\mathbb{E}
\left[
\left\|\mathbf{v}_\theta(\mathbf{A}_t^\tau,\mathbf{o}_t,\mathbf{h}_q, \tau) - (\boldsymbol{\epsilon}-\mathbf{A}_t)\right\|_2^2
\right]
\end{align}
where $(\mathbf{o}_t,\mathbf{A}_t)\sim D,\ \tau\sim p(\tau),\ \boldsymbol{\epsilon}\sim\mathcal{N}(0,\mathbf{I})$, and $\mathbf{h}_q$ is from backbone VLM.
To preserve pretrained multimodal knowledge and mitigate forgetting during large-scale robot pretraining, we freeze the full VLM by default and only train the learnable feature query, dataset adapters, and the DiT-based action head, following the common observation that retaining VLM priors is crucial for open-world reasoning and instruction grounding \cite{zhou2025chatvlaunifiedmultimodalunderstanding,hou2025diffusiontransformerpolicy}. 

In the task-specific fine-tuning stage, we adapt \svla to new tasks on \sarm using a small number of human demonstrations. Since fine-tuning solely on robot data can degrade multimodal understanding and spatial grounding, we augment fine-tuning with COCO \cite{coco2017}, VQA-v2 \cite{balancedvqav2}, and PixMo-Points \cite{pixmo}, following recent co-training practices \cite{pi0_5,zhou2025chatvla2visionlanguageactionmodelopenworld}. For a mixed batch consisting of robot samples $B_r$ and multimodal samples $B_m$, we jointly optimize the flow-matching control loss and the VLM next-token prediction loss:
\begin{align}
\mathcal{L}
=
\lambda_{\text{EM}}\cdot \mathcal{L}_{\text{EM}}(B_r)
+
\lambda_{\text{VLM}}\cdot \mathcal{L}_{\text{NTP}}(B_m),
\end{align}
where $\mathcal{L}$ is the total loss, $\lambda_{EM}$ and $\lambda_{VLM}$ are loss weights for embodied data and multi-modal data, and $\mathcal{L}_{\text{NTP}}$ is the standard autoregressive next-token prediction loss. We do not explicitly supervise chain-of-thought reasoning; nevertheless, when the VLM is fully frozen during pretraining, \svla retains the backbone’s conversational ability after post-training from simple experiments, suggesting that preserving VLM priors and multimodal co-training help maintain useful multimodal capabilities while learning embodied control.

\section{Hardware Design}


To adapt the robotic arm to human-centered desktop manipulation, we adopt a 6-DoF serial configuration as a practical abstraction of the human arm. We set 2 DoFs at the shoulder and 3 DoFs at the wrist, concentrating orientational dexterity at the distal end to better support diverse tabletop interactions. Our results show that this setting strikes a good balance between usability and cost. 

In particular, real-world manipulation requires the arm to exhibit high operability in task-relevant regions and long-term mechanical consistency, so as to ensure that the learned policy remains usable during deployment. We find that the arm's geometry structure, which are often described by the Modified Denavit-Hartenberg (MDH) parameters, are key factors determining these two characteristics. Then, we first model these two characteristics by the proposed two metrics: manipulation operability and endurance efficiency respectively and optimize these two metrics by searching for better MDH parameters. Details are as follows:

To facilitate the computation of these two metrics,  we first discretize the workspace $\mathcal{W} \subset \mathbb{R}^3$ into a voxel grid $\mathcal{V}$ of resolution $(N_X, N_Y, N_Z)$ with a width of 20mm for each voxel, and define $\nu(\cdot): \mathbb{R}^3\rightarrow(0,\dots,N_X-1)\times(0,\dots,N_Y-1)\times(0,\dots,N_Z-1)$ the voxel assignment function that maps a 3D point to its voxel index.


Then, we optimize the arm’s Modified Denavit-Hartenberg (MDH) parameters to better match the workspace and motion characteristics demanded by real-world tasks. In particular, real-world manipulation requires both high manipulation operability in task-relevant regions and  low mechanical wear for long-term consistency. 

Theseare strongly affected by link lengths and MDH geometry. We therefore formulate a multi-objective optimization problem with two quantitative objectives: Manipulation operability (\(F_1\)) and Endurance efficiency (\(F_2\)).

To compute these two metric, we discretize the workspace $\mathcal{W} \subset \mathbb{R}^3$ into a voxel grid $\mathcal{V}$ of resolution $(N_X, N_Y, N_Z)$ with a width of 20mm for each voxel, and define $\nu(\cdot): \mathbb{R}^3\rightarrow(0,\dots,N_X-1)\times(0,\dots,N_Y-1)\times(0,\dots,N_Z-1)$ the voxel assignment function that maps a 3D point to its voxel index.

Manipulation operability ($F_1$) is defined to quantify the richness of the kinematic solution space, which reflects how robustly the robotic arm can avoid singular configurations and motion interference when more feasible configurations are available. To compute this metric, we uniform sample joint configurations $q_i\in\mathbb{R}^6$ within the joint-limit hyper-rectangle $\mathcal{Q}=\prod_{d=1}^6[q_d^{\text{min}}, q_d^{\text{max}}]$, and map each sample to an end-effector position through forward kinetic function $\text{FK}(\cdot)$ with a total sample of $N$:

\begin{align}
    q_i \sim \mathcal{U}(\mathcal{Q}), x_i = \text{FK}(q_i) \in \mathbb{R}^3, i = 1, \dots, N
\end{align}

The number of feasible configurations falling into each voxel is then counted to obtain the manipulation operability density tensor $P_1$, which is defined as:

\begin{align}
    P_1[u,v,w] = \sum_{i=1}^N \mathbf{1}[\nu(x_i) = (u,v,w)]
\end{align}

where $\mathbf{1}[\cdot]$ is the indicator function and $(u,v,w)$ indexes the voxels. In parallel, teleoperation trajectories from task space are projected onto the same grid via $fk(\cdot)$ to get end-effector samples $y_i\in\mathbb{R}^3$, and the task density tensor \(P_2\) is defined as:

\begin{align}
    P_2[u,v,w] = \sum_{i=1}^N \mathbf{1}[\nu(y_i) = (u,v,w)]
\end{align}

Then we normalize the two tensors to $\tilde{P_1}$ and $\tilde{P_2}$ and constuct $F_1$:
\begin{align}
    F_1 = \frac{\sum_{(u,v,w) \in \Omega}\sqrt{\tilde{P_1}[u,v,w]\tilde{P_2}[u,v,w]}}{\sum_{(u,v,w) \in \Omega}\tilde{P_2}[u,v,w]}
\end{align}
with normalized term
\begin{align}
    \tilde{P_k}[u,v,w] = \frac{\log(1 + P_k[u,v,w]) - \min{\log(1+P_k)}}{\max{\log(1+P_k)} - \min{\log(1+P_k)}}     
\end{align}
where $\Omega = \{(u,v,w)\mid\tilde{P_2}[u,v,w] > 0\}$ denotes the task distribution, and $\sqrt{(\cdot)}$ implements a geometric-mean overlap that mitigrates over-penalization in low-density regions. Larger values of $F_1$ indicate that the robotic arm provides more redundant solutions in high-frequency task regions, and thus is better suited to complex manipulation.

\begin{figure}[H]
  \begin{center}
    \centerline{\includegraphics[width=\columnwidth]{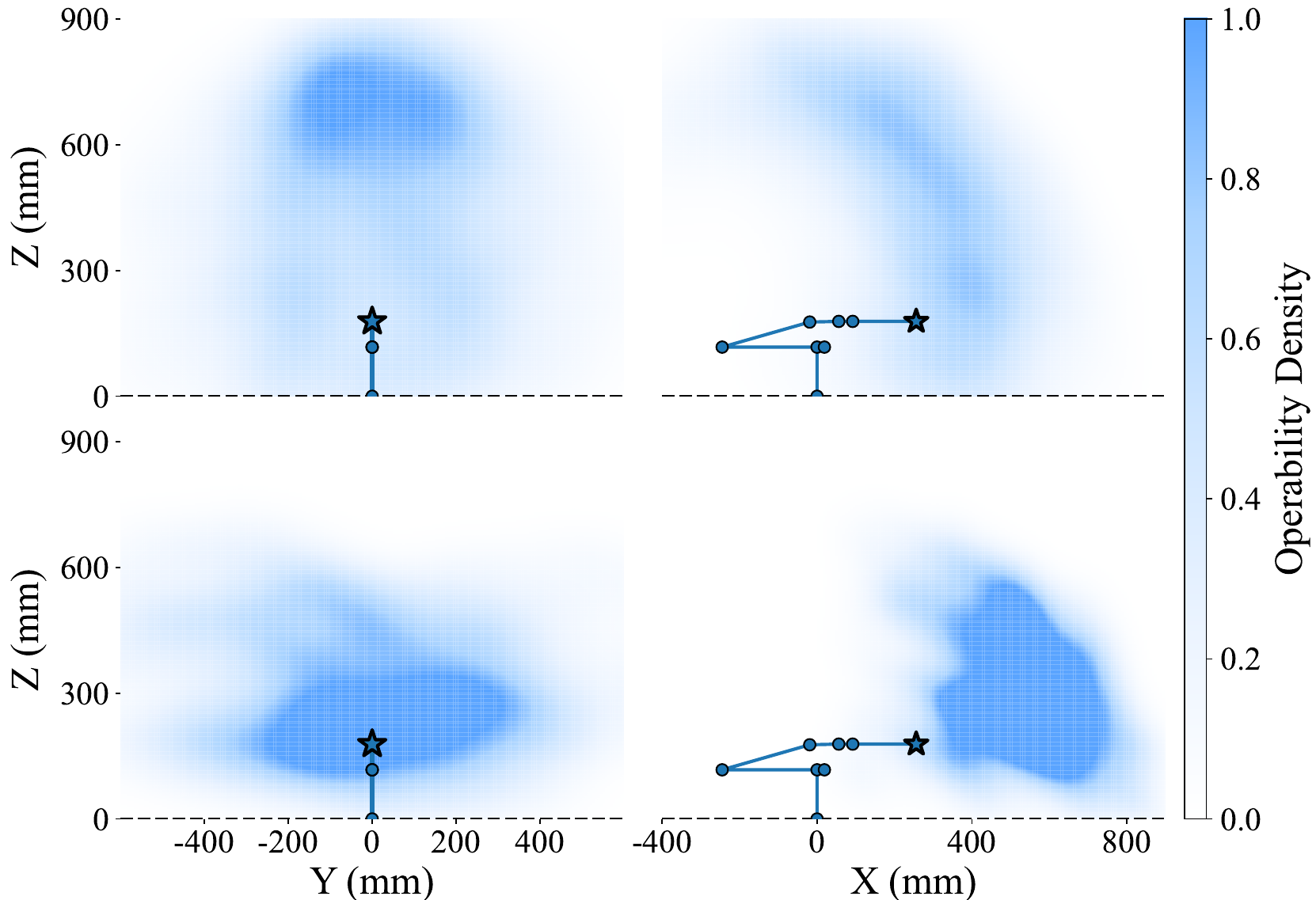}}
    \caption{
      Illustration of \(f_1\) calculation logic. Upper: Arm Operability. Lower: Task Operability.
    }
    \label{fig:overall_workspace}
  \end{center}
  \vskip -0.1in
\end{figure}




Endurance efficiency ($F_2$) characterizes the long-term operational costs of a robotic arm during operation. Sustained high costs intensity degrades the arm’s motion smoothness, exacerbates wear of its mechanical components, and thereby impairs its long-term operational reliability. The operational costs of a robotic arm can be modeled by the ``extent of effort'' required for the arm to maintain its current motion state. For each sample $q_i$, We compute $||\tau_i||$ as the normed joint torques required to maintain static equilibrium and the corresponding static power, and calculate the weighted sum at each voxel as $P_3$:

\begin{align}
    P_3[u,v,w] = \frac{\sum_{i=1}^N ||\tau_i|| \mathbf{1}[\nu(x_i) = (u,v,w)]}{\sum_{i=1}^N \mathbf{1}[\nu(x_i) = (u,v,w)]}
\end{align}

Since high-frequency task regions are more sensitive to sustained effort, the endurance efficiency metric $F_2$ is defined as

\begin{align}
    F_2 &= \frac{\sum_{(u,v,w) \in \Omega}\sqrt{\tilde{P_2}[u,v,w](1-\tilde{P_3}[u,v,w])}}{\sum_{(u,v,w) \in \Omega}\tilde{P_2}[u,v,w]}
\end{align}
where larger values of \(F_2\) indicate lower energy cost in task-dense regions and thus higher endurance efficiency.

Guided by standard adult anthropometric data \cite{gb10000_2023}, we jointly optimize manipulation operability and endurance efficiency with the NSGA-III algorithm \cite{deb2014nsgaiii}. Details of the optimization procedure and the resulting MDH parameters of \sarm are provided in Appendix~\ref{app:performance}.

 

  

\section{Control Design}

The primary objective of low-level control for a manipulator is to realize the target pose sequence produced by the VLA model as continuous and trackable joint commands. Since the VLA model outputs discrete key points as online-updated action chunks, chunk boundaries tend to induce pose discontinuities, consequently, discontinuities in joint velocity and acceleration. To improve execution stability, PID control is augmented with dynamics feedforward compensation(FF-PID), which attenuates nonlinear effects associated with velocity and acceleration and reduces the burden of feedback correction. To improve motion smoothness, a three-point Bézier action chunking interpolation scheme is further employed to continuousize discrete chunks and to generate smooth trajectories that satisfy joint actuation limits.

 \textbf{Arm Dynamic Model}. 
 Most industrial robotic arms rely on expensive precision manufacturing to reduce backlash and structural compliance, which in turn makes their dynamics easier to model accurately and supports high-performance model-based controllers such as MPC~\cite{dicarlo2018cheetah3_mpc}. 
 

\begin{figure*}[ht]
  \begin{center}
    \centerline{\includegraphics[width=\linewidth]{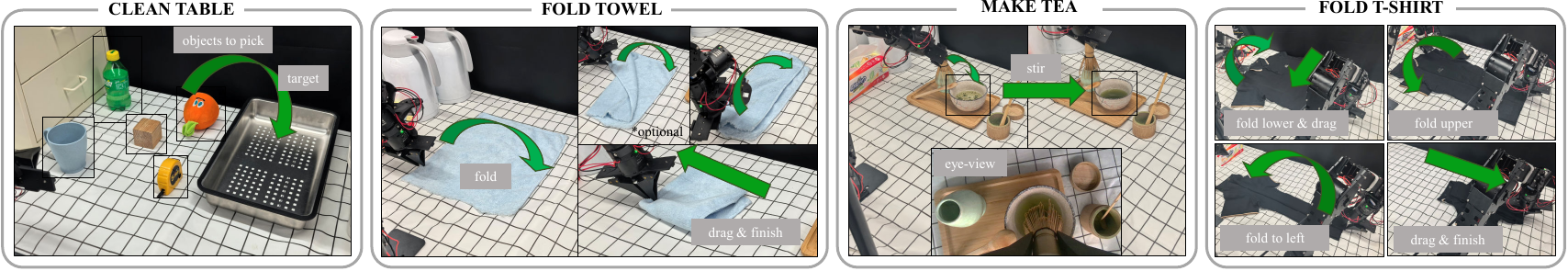}}
    \caption{
      Illustrations of tasks. We illustrate three real-world tasks involving from basic pick-place tasks to dual-arm cooperation on flexible soft cloth. These tasks are selected to verify the effectiveness of the robotic arm and model we use.
    }
    \label{fig:tasks}
  \end{center}
  \vskip -0.3in
\end{figure*}

To obtain reliable tracking performance under low-cost manufacturing conditions, we adopt a robust FF-PID control strategy guided by the Euler-Lagrange dynamics. The system model is established by defining the kinetic energy as $T(q, \dot{q}) = \frac{1}{2}\dot{q}^T M(q)\dot{q}$ and the potential energy as $V(q)$, then applying the Lagrange operator:

\begin{align}
\begin{cases}
\mathcal{L}(q, \dot{q}) = T(q, \dot{q}) - V(q), \\
\frac{d}{dt}\left(\frac{\partial \mathcal{L}}{\partial \dot{q}}\right) - \frac{\partial \mathcal{L}}{\partial q} = \tau.
\end{cases}
\end{align}

Where $q$ is defined as the position of the system. The dynamic equation of the system can be obtained as

\begin{align}
\tau = M(q)\ddot{q}+ C(q, \dot{q})\dot{q} + G(q)
\end{align}

Specifically, $M(q_d)$ is the inertia matrix derived from the kinetic energy term $T$, \(C(q,\dot q)\) is the Coriolis/centrifugal coupling matrix, which originates from the non-inertial nature of a rotating reference frame and becomes more pronounced as joint velocities increase, and $G(q_d) = \frac{\partial V}{\partial q}$ represents the gravity vector. 

To compensate for the known dynamic characteristics of the system and improve tracking accuracy, the torque \( \tau \) in dynamic equation of the system is taken as the model-based feedforward torque \( \tau_{ff} \). The overall control torque equation of the FF-PID strategy is expressed as:

\begin{align}
\begin{cases}
\tau = \tau_{ff} + \tau_{pid} + \tau_{fric}\text{sgn}(\dot{q}_d), \\
\tau_{ff} = M(q_d)\ddot{q}_d + C(q_d, \dot{q}_d)\dot{q}_d + G(q_d), \\
 \tau_{pid} = K_p e_q + K_i \int_0^t e_q(\varsigma)d\varsigma + K_d \dot{e}_q
\end{cases}
\end{align}

The feedback torque $\tau_{pid}$ governs the tracking error dynamics ($e = q_d - q$) to mitigate residual uncertainties, leveraging proportional $K_p$, integral $K_i$, and derivative $K_d$ gains to ensure the elimination of steady-state offsets and the attenuation of transient oscillations, respectively. In parallel, the friction compensation term, formulated as the product of the model magnitude $\tau_{fric}$ and the directional signum function $\text{sgn}(\dot{q}_d)$, serves to neutralize the system's inherent friction nonlinearities.

\textbf{Three-Point Rolling Bézier Action Chunking.} To generate a continuous and trackable joint-space reference across consecutive action chunks, we interpolate between adjacent keypoints using quintic polynomials. Since the VLA outputs only discrete joint positions \(q\in\mathbb{R}^6\), we estimate the endpoint derivatives required for interpolation online using a rolling three-point window: the current reached joint position \(q_c\), the next target joint position \(q_t\), and the subsequent target position \(q_n\). By finite-differencing these three 6D keypoints, we obtain the desired velocity and acceleration at \(q_t\).
\begin{align}
\dot{q}_t = \frac{q_n - q_c}{2\Delta t}, \qquad
\ddot{q}_t = \frac{q_n - 2q_t + q_c}{\Delta t^2},
\end{align}

To suppress torque spikes, we apply a jerk-bounded time-parameterization using a seven-phase S-curve profile, which enforces bounds on velocity, acceleration, and jerk while preserving smooth segment transitions. The full formulation and implementation details are provided in Appendix \ref{app:control}.

\section{Experiment}
\label{sec:setup}

We evaluate \sarm and \svla on four real-world manipulation tasks ranging from short-horizon pick-and-place to long-horizon dual-arm deformable-object manipulation, namely \textit{Clean Table}, \textit{Make Tea}, \textit{Fold Towel}, and \textit{Fold T-shirt} illustrated in Figure \ref{fig:tasks}, and full task definitions are provided in Appendix~\ref{app:tasks}. All comparisons are conducted under a unified evaluation protocol and are organized along two evaluation dimensions:
(i) \textbf{Hardware comparison}, by evaluating the self-developed algorithm \svla on \sarm as well as two widely used commercial 6-DoF robotic arms, ARX R5 and AgileX Piper, and collecting teleoperation datasets of matched size and fine-tune $\pi_0$ under identical training settings, after which the resulting policy is evaluated on the same task suite, and (ii)
\textbf{Model comparison}, by using \sarm as the evaluation robotic arm and fine-tune each baseline on the same dataset with the same training schedule, including ACT, earlier generalist policies Octo, OpenVLA-oft, and the $\pi$-series models $\pi_0$ and $\pi_{0.5}$.

\begin{table}[H]
  \centering
  \caption{Comprehensive performance comparison with existing platforms}
  \label{tab:performance}
  \footnotesize
  \begin{tabular}{lccc}
    \toprule
    \textbf{Metric} & \textbf{ARX R5} & \textbf{PIPER} & \textbf{\textit{Ours}} \\
    \midrule
    Manipulation operability & 0.546  & 0.179 & \textbf{0.547} \\
    Endurance Efficiency  & 0.529 & \textbf{0.846} & 0.567 \\
    Weight (kg)                     & 3.9  & 4.2  & \textbf{3.3} \\
    Material Cost (kUSD)   & 8.60 & 2.16 & \textbf{0.79} \\
    \bottomrule
  \end{tabular}
  \vspace{-0.8cm}
\end{table}
\vspace{0.2cm}

\subsection{Comparison with Other 6-dof Robotic Arms}
We first characterize the assembled \sarm by measuring its key hardware specifications, which are summarized in Table~\ref{tab:tech_specs} in Appendix~\ref{app:performance}.
Then, we compare \sarm against two widely used commercial 6-DoF arms, ARX R5 and AgileX Piper, to assess whether competitive desktop-manipulation capability can be achieved under a substantially lower cost budget; note that Piper adopts a distal two-coaxial joint design that alters its dexterity and workspace profile.

As summarized in Table~\ref{tab:performance}, \sarm attains a workspace volume comparable to the commercial arms while matching their torque margin, and it does so with lower weight and substantially reduced material cost.

\begin{table}[H]
  \caption{Task Success Rate of $\pi_0$ on Different Robotic Arms.}
  \label{tab:arm-comparison}
  \begin{center}
        \begin{tabular}{lccc}
          \toprule
          Task & \textbf{ARX R5} & \textbf{Piper} & \textbf{\textit{Ours}} \\
          \midrule
          Clean Table   & 0.88 & 0.86 & \textbf{0.92} \\
          Make Tea      & 0.40 & 0.40 & \textbf{0.60} \\
          Fold Towel    & 0.73 & \textbf{0.80} & 0.73 \\
          Fold T-shirt  & \textbf{0.83} & 0.50 & 0.75 \\
          \midrule
          Average       & 0.71 & 0.64 & \textbf{0.75} \\
          \bottomrule
        \end{tabular}
  \end{center}
  \vskip -0.2in
\end{table}

\begin{table*}[ht]
  \caption{Success Rate of Popular Models on Table Tasks collected by \sarm. \textbf{Clean Table} reports success rate of 5 in-domain objects. \textbf{Make Tea}, \textbf{Fold Towel} and \textbf{Fold T-shirt} reports single step success rates as well as final success rates. \textbf{Bold} marks best performance. \emph{Underline} means second best performance. Octo does not support multi-arm tasks.}
  \label{tab:model-comparison-table-total}
  \begin{center}
    \setlength{\tabcolsep}{3pt}
    \resizebox{\linewidth}{!}{%
        \begin{tabular}{l|ccccc>{\columncolor{gray!15}}c|ccc>{\columncolor{gray!15}}c|cccc>{\columncolor{gray!15}}c|ccccc>{\columncolor{gray!15}}c}
          \toprule
          \multirow{2}{*}{Model} & \multicolumn{6}{c|}{Clean Table} & \multicolumn{4}{c|}{Make Tea} & \multicolumn{5}{c|}{Fold Towel} & \multicolumn{6}{c}{Fold T-shirt (Dual-Arm)} \\
          & 1 & 2 & 3 & 4 & 5 & Avg. & 1 & 2 & 3 & Final & 1 & 2 & 3 & 4 & Final & 1 & 2 & 3 & 4 & 5 & Final \\
          \midrule
          ACT          & 0.8 & 1.0 & 1.0 & 0.7 & 0.1 & 0.72 & 1.0 & 0.7 & 0.86 & 0.6 & 1.0 & 0.6 & 0.69 & 0.56 & 0.33 & 0.08 & 0.0 & - & - & - & 0.0 \\
          Octo         & 0.6 & 0.1 & 0.3 & 0.0 & 0.0 & 0.2 & 0.0 & - & - & 0.0 & 0.13 & 0.0 & 0.0 & - & 0.0 & - & - & - & - & - & - \\
          OpenVLA-oft  & 1.0 & 0.9 & 1.0 & 0.5 & 0.0 & 0.68 & 0.2 & 0.0 & - & 0.0 & 0.6 & 0.4 & 0.67 & 1.0 & 0.27 & 0.25 & 0.66 & 0.0 & - & - & 0.0 \\
          $\pi_0$      & 0.8 & 1.0 & 0.8 & 1.0 & 1.0 & 0.92 & 0.9 & 0.78 & 0.86 & 0.6 & 0.93 & 0.8 & 0.85 & 1.0 & \emph{0.73} & 0.92 & 1.0 & 0.82 & 1.0 & 1.0 & \emph{0.75} \\
          $\pi_{0.5}$  & 1.0 & 1.0 & 0.8 & 1.0 & 1.0 & \textbf{0.96} & 1.0 & 0.8 & 1.0 & \textbf{0.8} & 0.93 & 0.8 & 1.0 & 0.92 & \textbf{0.8} & 0.92 & 1.0 & 0.91 & 1.0 & 1.0 & \textbf{0.83} \\
          \rowcolor{gray!20}
          Ours & 1.0 & 0.9 & 0.9 & 1.0 & 0.9 & \emph{0.94} & 0.9 & 0.78 & 1.0 & \emph{0.7} & 1.0 & 1.0 & 0.93 & 0.86 & \textbf{0.8} & 1.0 & 1.0 & 0.83 & 1.0 & 1.0 & \textbf{0.83}  \\
          \bottomrule
        \end{tabular}
    }
  \end{center}
  \vskip -0.2in
\end{table*}

We evaluate the usability of our robotic arm by running the same strong policy, $\pi_0$, on three different 6-DoF arms with same settings and comparing task success rates shown in Table \ref{tab:arm-comparison}.
Overall, $\pi_0$ achieves high average performance on our \sarm (0.75) and slightly higher than ARX R5 (0.71), suggesting that \sarm provides a stable and competitive platform for deploying modern VLA policies. In contrast, Piper attains a lower average success rate (0.64), with failures concentrated in grasp initiation and precise placement. We attribute this gap mainly to its distinct kinematic and actuation design, which yields different local dexterity and control response and makes precise grasp alignment harder under the same fine-tuning budget. These results justify using \sarm as our primary experimental platform and indicate that it can support high-quality manipulation performance comparable to a more expensive commercial alternative.

\subsection{Comparison with Other Models}
\label{sec:compare_to_models}

We then compare representative models on \sarm across four real-world tasks (\textit{Clean Table}, \textit{Make Tea}, \textit{Fold Towel}, and \textit{Fold T-shirt}), as summarized in \ref{tab:model-comparison-table-total}. For \textit{Clean Table}, we report the five objects in human manipluation data, and for the other tasks we report the final success rate as well as step-wise success rates. We fine-tune all baselines on the same in-domain dataset collected on \sarm with the same action representation and chunking setting, and use the same evaluation protocol across models. This setup isolates differences due to model capacity and action-generation mechanisms rather than task definitions or reward shaping.

Overall, earlier generalist policies (Octo and OpenVLA-oft) show limited effectiveness on our precision- and contact-rich tasks after fine-tuning, with failures typically occurring at grasp initiation or during contact transitions. ACT performs moderately on \textit{Clean Table} (0.72) and can solve a subset of rigid-object behaviors, but it degrades substantially on long-horizon deformable tasks, especially \textit{Fold T-shirt}, where compounding errors across sequential steps lead to unrecoverable states. This pattern is consistent with the general challenge of imitation learning under distribution shift, where small early errors can cascade in tasks requiring extended bimanual coordination.

In contrast, large-scale flow-based VLAs achieve strong and stable performance across the full suite. $\pi_0$ attains high success on \textit{Clean Table} (0.92) and remains competitive on \textit{Make Tea} and deformable manipulation, while $\pi_{0.5}$ further improves overall performance, achieving the best results on all the four tasks. In comparison, our \svla achieves moderately higher scores on the first three tasks comparing to $\pi_0$, and matches the best performance on \textit{Fold T-shirt} (0.83), despite being pretrained only on open-source datasets. This indicates that our design can effectively leverage the backbone VLM for spatial grounding and high-quality action generation even if the scale of our pretraining data is limited. The remaining gap to $\pi_{0.5}$ is plausibly driven by its substantially larger and higher-quality pretraining corpus.

\subsection{Ablation Study}
We conduct component ablations to quantify the contributions of our control stack and the \svla policy.


\begin{table}[H] 
\centering
\caption{Performance comparison and ablations of interpolation methods. Pos: Average Position Error (mrad). Vel: Average Velocity Error ($10^{-2}$ rad/s). Acc: Average Acceleration Error (rad/s$^2$). Tor: Average Torque Amplitude (N$\cdot$m). \textbf{Bold} marks best performance.}
\label{tab:interpolation_performance}
\setlength{\tabcolsep}{10pt}
\resizebox{1\linewidth}{!}{%
\begin{tabular}{@{}lcccc@{}}
\toprule
Method & Pos & Vel & Acc & Tor  \\
\midrule
\rowcolor{gray!15} Ours & \textbf{4.009} & 6.485 & \textbf{0.950} & \textbf{1.832} \\
\textit{\quad w/o Chunk-level smoothing} & 4.824 & 7.079 & 1.289 & 2.058 \\
\textit{\quad w/o Dynamics feedforward} & 6.369 & 19.925 & 86.306 & 8.630 \\
7-Segment S-curve & 9.062 & 7.835 & 1.316 & 2.473 \\
Jerk-limited (baseline) & 4.545 & \textbf{5.895} & 1.057 & 1.966 \\
\bottomrule
\end{tabular}
}
\vspace{-0.1in}
\end{table}

First, we evaluate our trajectory-execution stack (\textit{Ours}) against the industrial \textit{Jerk} baseline and a jerk-bounded \textit{7-stage} profile~\cite{robotoy_github}, using the metrics in Table~\ref{tab:interpolation_performance}. \textit{Ours} attains the best overall performance across all reported metrics. The average position error is $0.004009\,\text{rad}$, which is $13.3\%$ lower than \textit{Jerk} and $73.4\%$ lower than 7-stage; the maximum position error ($0.112538\,\text{rad}$) is $7.0\%$ lower than \textit{Jerk}. For motion smoothness, \textit{Ours} yields the lowest average acceleration error ($0.950438\,\text{rad/s}^2$), achieving a $10.1\%$ reduction relative to Jerk. The average torque amplitude is $1.832416\,\text{N}\cdot\text{m}$, which is $7.3\%$ lower than Jerk and $26.0\%$ lower than 7-stage. Furthermore, ablations on the Jerk-based pipeline show that removing upper-level VLA smoothing degrades all metrics, which increases position, velocity, acceleration, and torque errors by $6.1\%$, $20.1\%$, $21.9\%$, and $4.6\%$ relative to Jerk, respectively.
Additionally, removing dynamics compensation increases acceleration and torque errors by factors of approximately $70$ and $3.4$ relative to \textit{Jerk}, indicating that both modules are necessary for stable high-frequency execution.
\begin{table}[H]
  \caption{Task Success Rate of Different Architectures and training recipes of \svla. \textbf{Bold} marks best performance.}
  \label{tab:model-ablation}
  \begin{center}
    \resizebox{\linewidth}{!}{%
      \begin{small}
    \begin{tabular}{lccccc}
      \toprule
       \multirow{2}{*}{Model}  & \textbf{Clean} & \textbf{Make} & \textbf{Fold} & \textbf{Fold} & \multirow{2}{*}{\textbf{Avg.}} \\
        & \textbf{Table} & \textbf{Tea} & \textbf{Towel} & \textbf{T-shirt} &  \\
      \midrule
      \rowcolor{gray!20}
      \svla   & \textbf{0.94}  & \textbf{0.70}     & \textbf{0.80}   & \textbf{0.83}     & \textbf{0.82} \\
      \quad \textit{w/o dataset adapter}         & x     & x       & x       & x         & 0.00 \\
      \quad \textit{w/o multimodal data}           & 0.08  & 0.30     & 0.33    & 0.50       & 0.30 \\
      \quad \textit{w/ Qwen2.5-VL}      & 0.76  & 0.30     & 0.40    & 0.58      & 0.51 \\
      \quad \textit{w/ UMI datasets}          & 0.72  & 0.50     & 0.40    & \textbf{0.83}     & 0.61 \\
      \quad \textit{w/ relative action space}          & 0.90   & 0.40     & 0.53    & \textbf{0.83}     & 0.67 \\
      \quad \textit{w/ early VLM unfreezing}         & 0.78  & 0.40     & 0.47    & 0.42      & 0.52 \\
      \quad \textit{w/ late VLM unfreezing}         & 0.86  & 0.60     & 0.60    & \textbf{0.83}     & 0.72 \\
      \bottomrule
      \multicolumn{6}{l}{Note: x means failing to predict valid trajectories.} \\
    \end{tabular}%
    \end{small}
    }
  \end{center}
  \vskip -0.3in
\end{table}

We then conduct ablation to validate key design choices in \svla (Table~\ref{tab:model-ablation}). \svla achieves the best average success rate (0.80), and the largest degradations occur when weakening perception or cross-dataset alignment. Using previous version of Qwen-VL as VLM backbone drops the average score to 0.51, showing strong multimodal grounding is still crucial even with the same action head. Replacing dataset-specific adapters with a single shared adapter collapses performance to 0.0, underscoring the need to explicitly handle heterogeneous action/state conventions for scalable pretraining.

We also find that data composition, fine-tuning objectives, and adaptation strategy materially affect real-world performance. Adding a small portion of UMI data from community \cite{rayyan2025mv, zhu2025touch, liu2025vitamin, lin2024data, liu2024maniwav, ha2024umilegs, chi2024universalmanipulationinterfaceinthewild} helps long-horizon deformable manipulation but does not fully close the gap on precision tabletop tasks, suggesting diversity alone is insufficient without careful mixture alignment. Removing multimodal co-training during fine-tuning sharply reduces success to 0.30, supporting the role of auxiliary multimodal objectives in preserving spatial grounding and localization. For control and backbone adaptation, relative joint commands remain competitive (0.67), while early unfreezing hurts (0.52) but delayed unfreezing recovers much of the performance (0.72), indicating that keeping pretrained multimodal priors fixed early improves training stability and robustness.

\section{Conclusion}

Embodied intelligence is pushing AI from static perception toward real-world interaction, where sustained progress requires not only learning algorithms but also reproducible hardware and scalable data collection. We present \splatform, a fully open-source stack that integrates a low-cost, easy-to-build 6-DoF robotic arm (\sarm), a reliable control system, and an end-to-end VLA policy (\svla) trained via a reproducible two-stage recipe using only open-source robot and multimodal datasets. Experiments show that \sarm supports robust high-frequency execution and that \svla achieves competitive performance against mainstream VLA baselines. Beyond individual components, we release the complete pipeline, including mechanical designs and BOM, control codes, dataset conversion and processing tools, and training/inference scripts, to enable researchers to replicate, extend, and adapt the system for their own tasks and robots. 

Important challenges remain, particularly robustness to out-of-distribution objects and cross-embodiment generalization. Future work will improve \sarm with torque/impedance control, better sensing and calibration, and force-tactile grippers, and will strengthen \svla through larger open-source pretraining mixtures, improved data alignment, and objectives targeting OOD recovery. Crucially, by providing an end-to-end open platform, these directions can be pursued and validated by the community with minimal duplication of effort, enabling rapid iteration and cumulative progress on both hardware and learning systems. 

\section*{Impact Statement}

This paper presents work whose goal is to advance the field of Machine Learning. There are many potential societal consequences of our work, none which we feel must be specifically highlighted here.

\bibliographystyle{unsrtnat} 
\bibliography{siiarm} 

\end{multicols}
\newpage

\appendix

\counterwithin{equation}{section}
\counterwithin{figure}{section}
\counterwithin{table}{section}

\onecolumn

\section{Appendix}

\subsection{Performance Table Of the \sarm}
\label{app:performance}

\begin{table*}[h]
  \centering
  \caption{Modified D-H parameters of the robotic arm}
  \label{tab:dh_params}
  \begin{tabular}{@{}ccccc@{}}
        \toprule
        Link $i$ & Joint Angle $\theta_i$ (°) & Link Length $a_i$ (mm) & Link Offset $d_i$ (mm) & Joint Twist $\alpha_i$ (°) \\
        \midrule
        1 Base Rotation & 0 & 0 & 106.26 & 0 \\
        2 Shoulder Pitch & 180 & 19 & 0 & -90 \\
        3 Elbow Flexion & 180+$\theta$ & 269 & 0 & 180 \\
        4 Wrist Deflection & -$\theta$ & 236.12 & 0 & 0 \\
        5 Wrist Pitch & 90 & 80 & 0 & 90 \\
        6 End-Effector Rotation & 0 & 0 & 29 & 90 \\
        \bottomrule
    \end{tabular}
  \vskip 0.1in
  \caption*{\footnotesize
    Note: The parameter $\theta$ is a fixed angle shared by Link~3 and Link~4, with $\theta = 13.85^\circ$.
  }
  \vskip -0.2in
\end{table*}

\begin{table*}[h]
  \centering
  \caption{Technical specifications of the robotic arm}
  \setlength{\tabcolsep}{15pt}
  \label{tab:tech_specs}
  \footnotesize
  \begin{tabular}{@{}lc@{}}
    \toprule
    Quantity & Specification \\
    \midrule
    Active DOF                   & 6 \\
    Maximum Reach                & 636.7 mm \\
    Main Body Weight             & 3.3 kg \\
    Payload                      & 2 kg (at end-effector) \\
    Joint Repeatability          & $\leq \pm 0.03$ mm \\
    End-Effector Position Error  & $\leq 0.05$ mm \\
    Grasping Force Range         & 0.5--5 N (closed-loop) \\
    Communication Interface      & CAN \\
    Supply Voltage               & 24 V DC \\
    \bottomrule
  \end{tabular}
\end{table*}

\subsection{Data Collection Pipeline}

In \splatform, we implement multiple data collection methods that trade off cost, ease of deployment, and data quality. Although they differ in hardware requirements and interaction style, we align all outputs into a unified data format and reserve fields for straightforward conversion to other popular formats such as LeRobot and TFDS. 

\textbf{Drag Collection Mode} is integrated into \sarm's control library and allows users to directly drag the arm to complete task manipulations, requiring only a low-cost handle and two ferrules attached to the gripper. This mode is convenient for quickly validating reachability and end-to-end task feasibility with minimal external preparation, but it may occlude the collector’s view and the camera images, and the appearance of the human can introduce visual bias, increasing the burden of optional human-removal post-processing and the model’s robustness at inference. Direct manual operation can also reduce trajectory diversity, potentially hurting generalization. In this mode, for a trajectory of length $T$, we define
\begin{align}
        s_t &= (\tilde{p}_t, \tilde{v}_t, \tilde{e}_t, \tilde{c}_t), \\
        a_t &= \tilde{p}_{t+1} (1-\mathbf{1}_T(t)) + \tilde{p}_T \mathbf{1}_T(t), \nonumber \\
        r_t &= \mathbf{1}_T(t), \nonumber
\end{align}
where $s_t$ consists of sampled joint position $\Tilde{p}_t$, velocity $\Tilde{v}_t$, effort $\Tilde{e}_t$, and multi-camera pixels $\Tilde{c}_t$ read from motor sensors and cameras. Actions are taken as the next-step joint positions under the assumption that the arm can track $a_t$ within the sampling interval $\Delta t$; we use a default sampling frequency of 20\,Hz across all collection modes. Rewards are set to 1 only at the terminal step to indicate task completion, where $\mathbf{1}_{\hat{x}}(x)=1$ iff $x=\hat{x}$ and 0 otherwise.

\textbf{Multi-Arm Teleoperation Mode} uses a master--slave setup to collect demonstrations. The slave is a complete \sarm with a wrist camera, while the master can be another \sarm or a 3D-printed simplified \sarm model. We modify and integrate the GELLO system \cite{wu2024gellogenerallowcostintuitive} to synchronize and filter joint commands sampled from Dynamixel-based master motors to the slave arm. A 3D-printed master arm with servo motors costs around 250 USD, much cheaper of a full \sarm, where the servo motors account for most of the cost. Compared to drag collection, separating human operation from the robot workspace reduces human occlusion in the camera view, and because the slave executes the master’s commands, the resulting state distribution is more diverse and closer to deployment-time closed-loop control. However, the limited backdrivability and capability of servo motors restrict the master arm’s flexibility, making 3D-printed masters difficult to use in dual-arm teleoperation; a full dual-arm teleoperation setup may require four \sarm, increasing cost, although same-embodiment teleoperation remains a common choice. In teleoperation mode, we define
\begin{align}
        (p_t, v_t, e_t, c_t) &= f_{\Delta t}(a_{t-1}, p_{t-1}, v_{t-1}, e_{t-1}), \\
        s_t &= (\Tilde{p}_t, \Tilde{v}_t, \Tilde{e}_t, \Tilde{c}_t), \nonumber \\
        a_t &= \Tilde{p}_{t,m}, \nonumber \\
        r_t &= \mathbf{1}_T(t), \nonumber
\end{align}
where the real robot state evolves according to its dynamics $f_{\Delta t}(\cdot)$ and the sampled state is read from onboard sensors, while the action label is taken directly from the sampled master joint positions $\Tilde{p}_{t,m}$.

\textbf{VR Teleoperation Mode} replaces physical master arms with VR glasses and handheld controllers. The VR system provides controller poses (relative position and rotation), which we convert into joint-space control signals via inverse kinematics. VR is typically more expensive than 3D-printed master arms but often cheaper than adding two extra real arms, and it can control different robot types (including \sarm) without requiring dedicated master hardware, while also reducing space requirements and operator effort. In practice, non-negligible latency and pose noise from the device and network, indirect interaction with the robot workspace, and occasional inverse-kinematics failures can reduce trajectory smoothness and require additional operator training; limited battery capacity also constrains long continuous sessions. The state, reward, and system evolution are the same as teleoperation mode, but actions are computed by inverse kinematics:
\begin{align}
        (p_t, v_t, e_t, c_t) &= f_{\Delta t}(a_{t-1}, p_{t-1}, v_{t-1}, e_{t-1}), \\
        s_t &= (\Tilde{p}_t, \Tilde{v}_t, \Tilde{e}_t, \Tilde{c}_t), \nonumber \\
        a_t &= ik(\Tilde{q}_{t,VR}), \nonumber \\
        r_t &= \mathbf{1}_T(t), \nonumber
\end{align}
where $ik(\cdot)$ maps a VR pose to joint positions and can be extended to $ik(\Tilde{q}_{t,VR}, \Tilde{p}_{t-1}, \Tilde{v}_{t-1}, \Tilde{e}_{t-1})$ for post-processing such as joint-limit enforcement, filtering, and smoothing.

Finally, we integrate \textbf{UMI} \cite{chi2024universalmanipulationinterfaceinthewild}, which uses visual SLAM-based tracking instead of VR controllers for pose estimation and data collection. We design a parallel gripper to fit UMI into \sarm and replace its camera with a standard depth camera, making the resulting data closer to our other collection modes and easier to co-pretrain with additional sources. UMI does not require access to real robot arms during collection and is therefore the most flexible option for gathering large-scale pretraining data in diverse environments. However, due to limitations of visual SLAM, it exhibits similar pose accuracy issues and inverse-kinematics failures, making it less suitable for collecting high-precision fine-tuning data for real manipulation. Its state, action, and reward definitions are
\begin{align}
        \Tilde{p}_t &= ik(\Tilde{q}_{t,UMI}), \\
        \Tilde{v}_t &= \frac{\Tilde{p}_{t+1} - \Tilde{p}_t}{\Delta t} \times (1 - \mathbf{1}_T(t)), \nonumber \\
        \Tilde{e}_t &= 0, \nonumber \\
        s_t &= (\Tilde{p}_t, \Tilde{v}_t, \Tilde{e}_t, \Tilde{c}_t), \nonumber \\
        a_t &= \Tilde{p}_{t+1} (1-\mathbf{1}_T(t)) + \Tilde{p}_T \mathbf{1}_T(t), \nonumber \\
        r_t &= \mathbf{1}_T(t), \nonumber
\end{align}
where the UMI pose $\Tilde{q}_{t,UMI}$ is mapped to virtual joint positions, velocity is estimated by finite differences and set to 0 at the terminal step, effort is set to 0 since it is not directly measurable, and actions follow the next-step joint position labeling as in drag collection mode.

\subsection{Processes of the Arm Design}
We define two voxel-based objectives to make the arm geometry explicitly match the spatial distribution of human-centered desktop tasks. The key idea is that desktop manipulation does not use the entire reachable workspace uniformly: tasks concentrate on a small subset of regions (e.g., above the table and within comfortable reach). Therefore, instead of optimizing global workspace volume, we optimize (i) how kinematically redundant the arm is in task-dense regions and (ii) how costly it is to hold/operate in those regions.

\textbf{ \(F1\) (Manipulation Operability).}
We voxelize the 3D Cartesian workspace into a uniform grid with voxel edge length $\Delta=20$\,mm, and index all voxels by $(u,v,w)\in\{0,\dots,N_X\!-\!1\}\times\{0,\dots,N_Y\!-\!1\}\times\{0,\dots,N_Z\!-\!1\}$. We define $P_1[u,v,w]$ as the operability density of voxel $(u,v,w)$, i.e., the number of feasible joint configurations (within joint limits and collision-free) whose end-effector position falls into this voxel. This density reflects the richness of the kinematic solution space in the local region: larger $P_1[u,v,w]$ means more available configurations and thus more flexibility to avoid singularities, self-collisions, and motion interference during task execution. Concretely, we uniformly sample joint configurations $q_i\in\mathcal{Q}$, compute end-effector positions $x_i=fk(q_i;\theta)$, and accumulate them into voxels:
\begin{align}
P_1[u,v,w]=\sum_{i=1}^{N}\mathbb{I}\!\left[\nu(x_i)=(u,v,w)\right],
\end{align}
where $\mathbb{I}[\cdot]$ is the indicator function, $N$ is the number of workspace samples, and $\nu(\cdot)$ denotes the voxel indexing (mapping) function. We define $P_2[u,v,w]$ as the task density of voxel $(u,v,w)$, i.e., how frequently the task visits this region, constructed by voxel counting over task-space samples $\{y_j\}_{j=1}^{N}$:
\begin{align}
P_2[u,v,w]=\sum_{j=1}^{N}\mathbb{I}\!\left[\nu(y_j)=(u,v,w)\right].
\end{align}
Since $P_1$ and $P_2$ are generated by different processes (workspace sampling vs. task sampling) and thus have different numeric scales, we independently normalize each voxel field to $[0,1]$:
\begin{align}
\tilde{P}[u,v,w]=\frac{P[u,v,w]-\min_{(a,b,c)}P[a,b,c]}{\max_{(a,b,c)}P[a,b,c]-\min_{(a,b,c)}P[a,b,c]+\epsilon},\qquad \epsilon=10^{-8},
\end{align}
and denote $\tilde{P}_1=\widetilde{P_1}$ and $\tilde{P}_2=\widetilde{P_2}$. We then measure task-weighted operability by combining them voxel-wise: voxels that are frequently used by the task should contribute more to the objective. Then, we have
\begin{align}
\mathbf{f1}=\sum_{u=0}^{N_X-1}\sum_{v=0}^{N_Y-1}\sum_{w=0}^{N_Z-1}\sqrt{\tilde{P}_1[u,v,w]\,\tilde{P}_2[u,v,w]}.
\end{align}
This objective is large when the arm provides high solution richness in regions that are frequently visited by the task distribution.

\textbf{ \(F2\) (Endurance Efficiency).}
Beyond manipulation operability, desktop manipulation also benefits from designs energetically favorable in task-dense regions. For low-cost servo-driven robotic arms, high required holding torque typically correlates with increased current draw, excessive heating, and diminished control margin. We therefore construct a voxel-wise endurance cost field \(P_3[u,v,w]\) from a static gravity-torque proxy, and further convert it into an efficiency field (higher is better) to align the optimization direction with \(F1\).
For each valid workspace sample \(q_i\) (reaching \(x_i\)), we compute the gravity torque vector \(\tau_g(q_i)\in\mathbb{R}^{6}\) and map it to a scalar effort cost:
\begin{align}
e(q_i)=\sum_{j=1}^{6}\left|\tau_{g,j}(q_i)\right|.
\end{align}
We aggregate this sample-level cost into a voxel-level average cost:
\begin{align}
P_3[u,v,w]=
\frac{\sum_{i=1}^{N} e(q_i)\cdot \mathbb{I}\!\left[\nu(x_i)=(u,v,w)\right]}
{\sum_{i=1}^{N}\mathbb{I}\!\left[\nu(x_i)=(u,v,w)\right]+\epsilon}.
\end{align}
and normalize it independently to \([0,1]\), denoted as $\tilde{P}_3$, where larger \(\tilde{P}_3[u,v,w]\) indicates higher endurance cost in voxel \((u,v,w)\). We then convert this cost into an efficiency field (higher is better):
\begin{align}
P_{3\mathrm{eff}}[u,v,w]=1-\tilde{P}_3[u,v,w].
\end{align}
Finally, we compute task-weighted energy efficiency by multiplying with the task density and summing over voxels:
\begin{align}
\mathbf{f2}
=\sum_{u=0}^{N_X-1}\sum_{v=0}^{N_Y-1}\sum_{w=0}^{N_Z-1}
\sqrt{\tilde{P}_2[u,v,w]\;P_{3\mathrm{eff}}[u,v,w]}
=
\sum_{u=0}^{N_X-1}\sum_{v=0}^{N_Y-1}\sum_{w=0}^{N_Z-1}
\sqrt{\tilde{P}_2[u,v,w]\;\left(1-\tilde{P}_3[u,v,w]\right)}.
\end{align}
In implementation, we normalize the voxel-wise summation by the total task mass to obtain scale-stable objectives in \([0,1]\). Concretely, we compute
\begin{align}
\mathbf{F1}
&=
\frac{
\sum\limits_{u=0}^{N_X-1}\sum\limits_{v=0}^{N_Y-1}\sum\limits_{w=0}^{N_Z-1}
\sqrt{\tilde{P}_1[u,v,w]\;\tilde{P}_2[u,v,w]}
}{
\sum\limits_{u=0}^{N_X-1}\sum\limits_{v=0}^{N_Y-1}\sum\limits_{w=0}^{N_Z-1}
\tilde{P}_2[u,v,w]+\epsilon
},
\\
\mathbf{F2}
&=
\frac{
\sum\limits_{u=0}^{N_X-1}\sum\limits_{v=0}^{N_Y-1}\sum\limits_{w=0}^{N_Z-1}
\sqrt{\tilde{P}_2[u,v,w]\;P_{3\mathrm{eff}}[u,v,w]}
}{
\sum\limits_{u=0}^{N_X-1}\sum\limits_{v=0}^{N_Y-1}\sum\limits_{w=0}^{N_Z-1}
\tilde{P}_2[u,v,w]+\epsilon
}
\\
&=
\frac{
\sum\limits_{u=0}^{N_X-1}\sum\limits_{v=0}^{N_Y-1}\sum\limits_{w=0}^{N_Z-1}
\sqrt{\tilde{P}_2[u,v,w]\;\left(1-\tilde{P}_3[u,v,w]\right)}
}{
\sum\limits_{u=0}^{N_X-1}\sum\limits_{v=0}^{N_Y-1}\sum\limits_{w=0}^{N_Z-1}
\tilde{P}_2[u,v,w]+\epsilon
},
\end{align}
where \(\epsilon=10^{-8}\) is a numerical constant. This task-normalization makes $F1$, $F2$  comparable across different sampling sizes and voxel resolutions while preserving the same task-weighted optimization objective.

Guided by standard adult anthropometric data, we formulate a two-objective optimization over the arm MDH geometry to jointly maximize manipulation operability (\textbf{\(F1\)}) and endurance efficiency (\textbf{\(F2)\)}). We optimize the decision vector $\mathbf{x}=[a_2,a_3,l_{\mathrm{gripper}}]^\top$, where $a_2\in[240,300]$\,mm and $a_3\in[210,260]$\,mm correspond to the human upper arm and forearm lengths, and $l_{\mathrm{gripper}}\in[170,200]$\,mm controls the effective gripper length and grasping posture. All remaining MDH parameters (including link twists $\alpha_i$) are fixed to simplify low-cost manufacturing and preserve kinematic stability.

We solve this multi-objective problem using NSGA-III. The initial population is generated by Latin Hypercube Sampling (LHS) with population size $N=\mathbf{200}$.  We use simulated binary crossover (SBX) with $p_c=\mathbf{0.9}$ and $\eta_c=\mathbf{20}$, The final output is a Pareto set $\mathcal{P}$.

To select a single design from $\mathcal{P}$, we compute entropy-based objective weights $(\omega_1,\omega_2)$ from an LHS sample set of $n=\mathbf{200}$ candidates. Let $X=(x_{ij})\in\mathbb{R}^{n\times 2}$ be the objective matrix with $x_{i1}=\mathbf{F1}(\mathbf{x}_i)$ and $x_{i2}=\mathbf{F2}(\mathbf{x}_i)$. We apply min--max normalization
\begin{align}
r_{ij}=\frac{x_{ij}-\min_i x_{ij}}{\max_i x_{ij}-\min_i x_{ij}+\epsilon},\qquad \epsilon=10^{-8},
\end{align}
compute
\begin{align}
p_{ij}=\frac{r_{ij}+\varepsilon}{\sum_{i=1}^{n}(r_{ij}+\varepsilon)},\qquad \varepsilon=10^{-6},
\end{align}
and the entropy
\begin{align}
e_j=-\frac{1}{\ln n}\sum_{i=1}^{n}p_{ij}\ln p_{ij}.
\end{align}
Finally,
\begin{align}
\omega_j=\frac{1-e_j}{\sum_{k=1}^{2}(1-e_k)},\qquad j\in\{1,2\}.
\end{align}
In our experiments, we obtain $\omega_1=\mathbf{0.43}$ and $\omega_2=\mathbf{0.57}$, and select the final design by maximizing
\begin{align}
S(\mathbf{x})=\omega_1\,\mathbf{F1}(\mathbf{x})+\omega_2\,\mathbf{F2}(\mathbf{x}),\qquad
\mathbf{x}^\star=\arg\max_{\mathbf{x}\in\mathcal{P}} S(\mathbf{x}).
\end{align}
The resulting MDH parameters of \sarm are reported in Table~\ref{tab:dh_params}.

\subsection{Gripper Design}

\begin{figure*}[ht]
  \vskip 0.2in
  \begin{center}
    \centerline{\includegraphics[width=\linewidth]{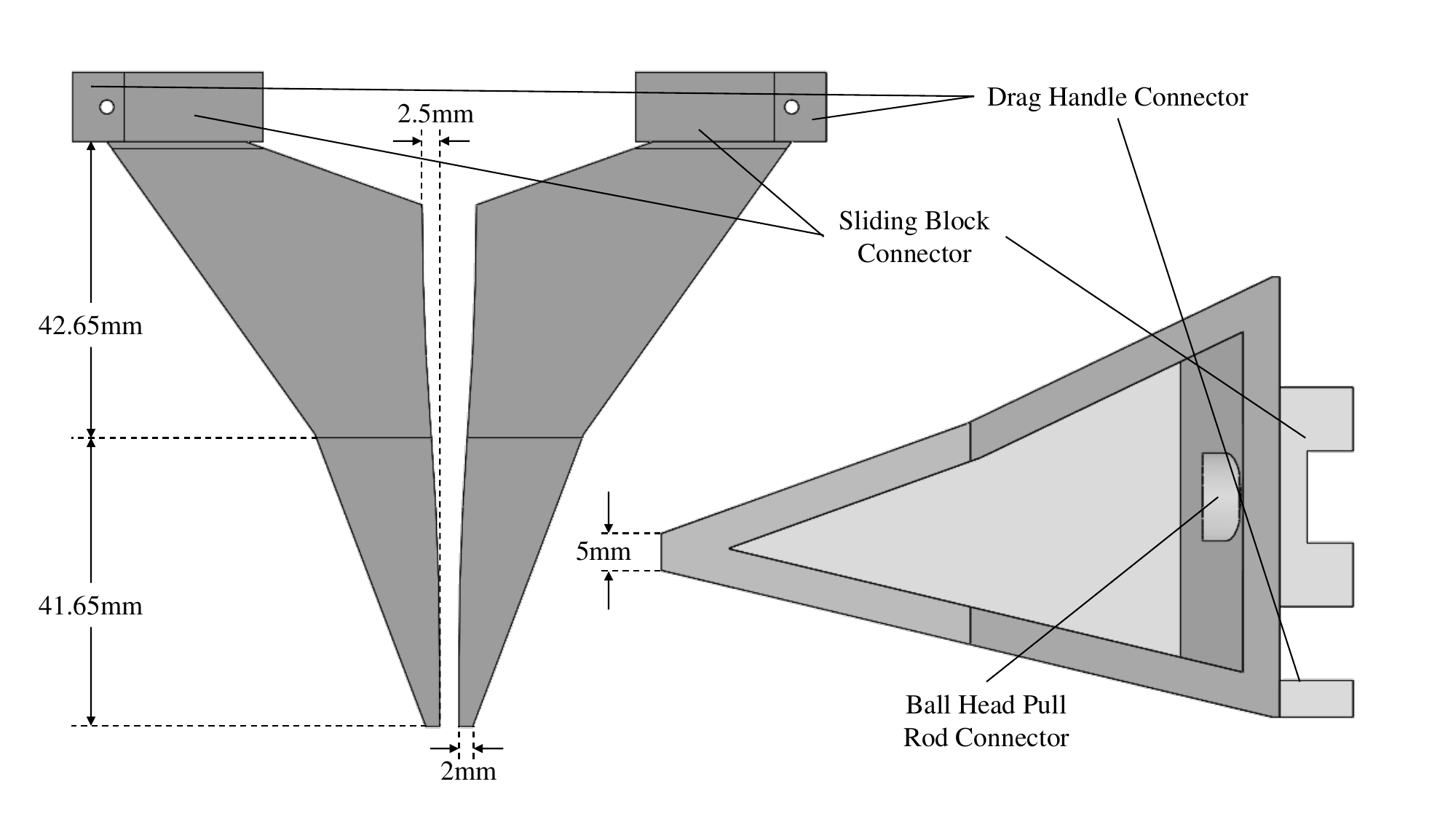}}
    \caption{
      Gripper Design Details. \textit{Left}: top view of gripper. \textit{Right}: right view of gripper. Each gripper has one connector for drag handle, one for sliding block and one for the ball head pull rod. The tip of the gripper is sharpened to have only an area of 5mm $\times$ 2mm, and the surface is designed to be s-curved to focus the gripping force to the end of the gripper. The front half of the gripper is also thinner to reach into small openings more easily.
    }
    \label{fig:gripper}
  \end{center}
\end{figure*}

\begin{figure*}[ht]
  \vskip 0.2in
  \begin{center}
    \centerline{\includegraphics[width=\linewidth]{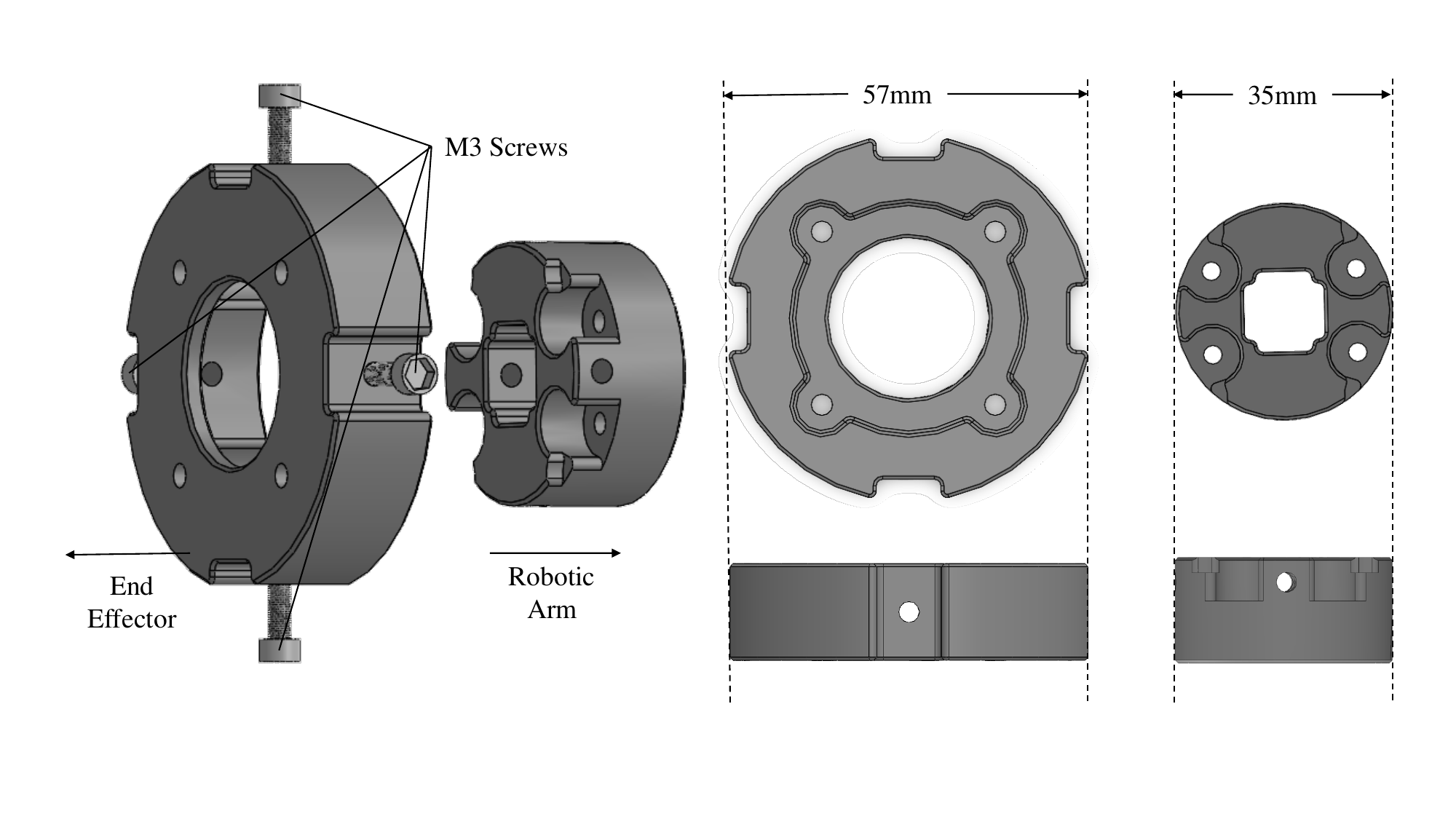}}
    \caption{
      Gripper Interface Design Details. \textit{Left}: Overview of the interface. \textit{Middle}: front and side view of the interface connected to the end effector. \textit{Right}: front and side view of the interface connected to the robotic arm. We use four M3 screws to secure the two parts together. Both parts can be substituted to connect to other flanges of end effectors.
    }
    \label{fig:gripper_interface}
  \end{center}
\end{figure*}

We develop a two-finger parallel gripper as the default end-effector of \sarm, illustrated in \cref{fig:gripper}. As another important part on the robotic arm, we need the gripper to interact with the external world efficiently and can complete a vast range of common tasks. Besides, the gripper should be flexible and easy to substitute for other types of grippers, like dexterous hands. Based on actual usage scenarios, we define four main requirements for our gripper:
\begin{enumerate}
    \item \textbf{Generalization}: The gripper should be capable of tightly gripping various kind of objects of different materials.
    \item \textbf{Wide Range}: The gripper should have a large gripper width for large objects, or can be easily expanded for a larger width on demand.
    \item \textbf{Lightweight}: The gripper should have a small weight to leave loading to the gripped object and maintain control accuracy.
    \item \textbf{Decoupling}: The gripper should be decoupled with the arm's control, with fewest parameters like gripper mass and max gripper width.
\end{enumerate}

We first choose a parallel gripper design instead of a ball-screw-and-rod design. These two grippers have different motion modes and operating principles. With a parallel gripper, the two fingers translate in a straight line within the same plane, which keeps the grasp point unchanged across different gripper widths and thus improves stability and accuracy during grasping. However, it requires fixed slide rails, leading to a base whose width is larger than the finger width, which may prevent the robotic arm from reaching into containers with small openings such as packing boxes. By contrast, a rod gripper can save horizontal space, but its grasp point varies with the gripper width. It also requires precise metal processing, which increases the overall weight and the complexity of the conversion between gripper width and motor readings. When expanding to a larger gripper width, a parallel gripper only needs two longer slide rails, whereas a rod gripper must be redesigned and remanufactured. Based on \textit{requirement 2 and 3}, we ultimately choose the parallel gripper as our default end effector. We use transducers and ball-head pull rods to convert the motor’s rotational motion into the gripper’s translational motion. We use metal parts only where needed to strengthen structural integrity and prevent the gripper from breaking under excessive forces during motion. The gripper base and fingers are 3D-printed for flexible manufacturing and easy modification.

We then carefully design the gripper geometry. Beyond gripping regular rigid objects, we enhance its ability to grasp soft flexible cloth and large round bottles, while also improving its limitation in reaching into small openings. We design a sharp, inwardly protruding fingertip so that, when the gripper closes, only the sharp tips make tight contact. This reduces the contact area and concentrates grasping pressure on soft flexible cloth. In addition, we reduce the height of the front half of the gripper so that it can insert deeper into small openings. For the finger surface, we adopt an S-curve profile to better fit large cylindrical bottles. Beyond shaping the contact surface, we perform shell extraction on the outer structure to further reduce the gripper’s weight and attach rubber anti-slip tapes on the finger surfaces to increase friction against objects. These detailed designs improve the success rate and robustness of grasping across different scenarios, satisfying \textit{requirement 1} for objects of various materials and forms.

For gripper substitution scenarios, we design simple interfaces so that the robotic arm can quickly adapt to more tasks, as shown in \ref{fig:gripper_interface}. We use four M3 screws arranged in a cross pattern to mount the gripper and its motor, and the motor is connected to the robotic arm’s last joint motor via a series-connected CAN bus and power supply using a unified XT30 2+2 wire. With an appropriately sized adapter, any end effector can be attached to the flange and its wire can be connected either to the robotic arm’s last motor using the same communication and power protocol, or directly to the computer. We also use screws to connect grippers to the ball-head pull rods and sliding blocks. These grippers can be readily redesigned and remanufactured via 3D printing and replaced to achieve longer gripper depth or more suitable shapes, and the gripper base can be expanded for a larger gripper width. Under \textit{requirement 4}, these detachable components provide \sarm with more possibilities for performing a wider range of tasks.

\subsection{Control Design}
\label{app:control}

\textbf{Three-Point Rolling Bézier Action Chunking.}We propose a three-point Bézier action-chunking interpolation method that generates smooth trajectories under physical limits by explicitly imposing velocity and acceleration constraints at key points. In implementation, transitional points are inserted between adjacent chunks and the motion interval is subdivided, which attenuates inter-chunk discontinuities. Within each sub-interval, quintic polynomial interpolation is employed so that position, velocity, and acceleration boundary constraints can be satisfied simultaneously, thereby ensuring smoothness at the velocity and acceleration levels.

Since an action chunk does not provide the endpoint velocity and acceleration required for interpolation, a rolling three-point finite-difference scheme is introduced to estimate the motion state at key points online, and the resulting estimates are used as boundary conditions for the quintic polynomials. Let \(q_c\), \(q_t\), and \(q_n\) denote the predecessor, target, and successor points in the rolling window, respectively, and let \(\Delta t\) be the time step. The velocity and acceleration at \(q_t\) are estimated as
\begin{align}
\dot{q}_t = \frac{q_n - q_c}{2\Delta t}, \qquad
\ddot{q}_t = \frac{q_n - 2q_t + q_c}{\Delta t^2},
\end{align}
where \(\dot{q}_t\) and \(\ddot{q}_t\) denote the desired velocity and acceleration at \(q_t\). These second-order finite-difference estimates supply the boundary parameters required by quintic interpolation and improve inter-segment continuity.

To suppress torque spikes and impact-inducing transients, jerk is constrained so that acceleration changes are smoothed. Building on the smoothly interpolated trajectory, the motion is partitioned into seven phases, comprising ramp-up, constant-acceleration, ramp-down, constant-velocity, and the corresponding deceleration stages.

Let $T$ denote the total duration of a smooth segment, and let $\alpha_i$ $(i=1,\dots,7)$ denote the normalized duration of the $i$-th phase, satisfying $\alpha_i > 0$ and $\sum_{i=1}^7 \alpha_i = 1$. Define the cumulative phase boundaries
\begin{align}
S_k = \sum_{i=1}^k \alpha_i,\quad k=1,\dots,7,
\end{align}
with $S_0 = 0$ and $S_7 = 1$. The jerk profile $j(t)$ is then specified as
\begin{align}
j(t)=
\begin{cases}
j_1 & (0 \leq t \leq S_1 T) \\
0 & (S_k T \leq t \leq S_{k+1} T),\ k=1,3,5 \\
-j_1 & (S_2 T \leq t \leq S_3 T) \\
j_2 & (S_4 T \leq t \leq S_5 T) \\
-j_2 & (S_6 T \leq t \leq T)
\end{cases}
\end{align}

where $j_1$ and $j_2$ are the nonzero jerk magnitudes used in the acceleration and deceleration phases, respectively.

\subsection{Tasks}
\label{app:tasks}
We selected 4 different tasks to test our \splatform, which covers a range from basic pick-place task to dual-arm flexible object manipulation. These tasks are specifically selected to compare the performance of different robotic arms and models in difficult scenarios.

 In \textbf{\textit{Clean Table}}, the robot needs to pick the object on the table to the metal plate. In the manipulation dataset we prepare five different objects including wooden cube, carrot toy, blue cup, yellow tape measure, and green bottle, to test the robotic arm's gripping ability for regular, soft, thin, heavy, round and large objects, respectively. We also prepare three extra objects that are not collected in fine-tuning data during testing. Although the task's operating procedure is relatively simple, this task can still verify that the robotic arm and its policy have basic operational capabilities, and also ensure that the robotic arm can handle objects with different characteristics, and that the policy has a certain degree of generalization ability.

 In \textbf{\textit{Make Tea}}, the robot needs to pick up the bamboo whisk to whisk the matcha powder in the bowl until it's evenly mixed, then places the whisk back onto the whisk stand, indicating the completion of the task. To complete the task perfectly, the robot needs to accurately observe the changes in the matcha bowl, ensuring that it captures the information that the matcha powder has been evenly mixed into the bowl, neither removing the whisk too early nor continuing to stir infinitely. This task requires both the model's ability to observe details and the robotic arm's ability to perform precise manipulations. At the same time, it can also serve as a demonstration that can be applied to real-world situations, verifying the operational capabilities of the robotic arm model in real-world scenarios and promoting its practical applicability.

 In \textbf{\textit{Fold Towel}}, the robot needs to fold the light blue towel in front of it into four folds, and then move it to the left of the table. As a single-arm task, the robot needs to grasp the left edge of the towel and fold it to the right twice to complete the first fold, and then fold the bottom edge of the towel to the top to complete the second fold before dragging the towel to the left. This task requires the precision of both the model and the robot to grip the edge of the towel close to the table and the generalization ability to handle positional changes of the towel during manipulation. Many tasks involving manipulating flexible objects themselves usually use two robotic arms to prevent the flexible objects from sliding excessively across the table, so this single-arm task also requires the stability of the arm and the model.

In \textbf{\textit{Fold T-shirt}}, the robot needs to fold the T-shirt on the table in predefined steps. When a T-shirt is placed horizontally on the table, the robot needs to use its two arms concurrently to fold the sleeve near the robot into the body, drag the T-shirt closer, and then fold the sleeve on the other side back. Then, the robot needs to fold the bottom of the T-shirt on the right side to its collar on the left side. Finally, the robot drags the folded T-shirt to the right corner of the table and finishes the task. In this task, the model needs to deal with multiple status of the T-shirt, operate both robotic arms at the same time, and also grip the cloth precisely. This is a long-horizon task with complex interactions with flexible objects and cooperation between two robotic arms. Besides, different from single arm tasks, we use VR teleoperating systems to manipulate on this task to demonstrate the adaptability of robotic arms and policies in the face of manipulation data from different sources.



\end{document}